\pdfoutput=1

\documentclass[11pt]{article}

\usepackage{EMNLP2022}
\usepackage{graphicx}
\usepackage{xcolor}
\usepackage{array}
\usepackage{caption}
\usepackage{subcaption}

\usepackage{times}
\usepackage{latexsym}
\usepackage{tabularx}
\usepackage{multirow}

\usepackage[T1]{fontenc}

\usepackage[utf8]{inputenc}

\usepackage{microtype}

\usepackage{inconsolata}
\usepackage{verbatim} 
\title{Exploring Robustness of Prefix Tuning in Noisy Data: A Case Study in Financial Sentiment Analysis}

\author{Sudhandar Balakrishnan \and Yihao Fang \and Xiaodan Zhu \\
        Department of Electrical and Computer Engineering \& Ingenuity Labs Research Institute \\
        Queen's University \\ 
        \{sudhandar.balakrishnan, yihao.fang, xiaodan.zhu\}@queensu.ca}

\begin{document}
\maketitle
\begin{abstract}

The invention of transformer-based models such as BERT, GPT, and RoBERTa has enabled researchers and financial companies to finetune these powerful models and use them in different downstream tasks to achieve state-of-the-art performance. Recently, a lightweight alternative (approximately 0.1\% - 3\% of the original model parameters) to fine-tuning, known as prefix tuning has been introduced. This method freezes the model parameters and only updates the prefix to achieve performance comparable to full fine-tuning. Prefix tuning enables researchers and financial practitioners to achieve similar results with much fewer parameters. In this paper, we explore the robustness of prefix tuning when facing noisy data. Our experiments demonstrate that fine-tuning is more robust to noise than prefix tuning---the latter method faces a significant decrease in performance on most corrupted data sets with increasing noise levels. Furthermore, prefix tuning has high variances in the F1 scores compared to fine-tuning in many corruption methods. We strongly advocate that caution should be carefully taken when applying the state-of-the-art prefix tuning method to noisy data. 

\end{abstract}

\section{Introduction}

The transformer architecture \citep{vaswani2017attention} has given rise to several powerful language models such as BERT \citep{devlin2018bert} and GPT \citep{radford2018improving}. These models are trained on large text corpora and the pre-trained models can be used on different downstream tasks by finetuning these models, which refers to the process of updating the weights of the pre-trained model to adapt to the downstream task and the associated dataset. This approach is critical in achieving state-of-the-art results in many downstream tasks. However, these fine-tuned language models are large in size and the deployment of these models in production to solve real-world problems becomes difficult due to the memory requirement, constraining the deployment of models in many real-life financial applications. Given that it is anticipated that model sizes will continue to rise, this will become more serious.

\citet{li2021prefix} introduced a lightweight alternative to finetuning known as prefix tuning. The authors freeze the model parameters of GPT-2 \citep{radford2019language} and use a task-specific vector to tune the model for natural language generation. This method achieves comparable performance with finetuning and uses approximately 0.1\% - 3\% of the original model parameters. This method will enable the use of pre-trained language models for many industrial applications. 

In the financial sector, natural language processing has a wide variety of applications ranging from building a chatbot to interact with customers \citep{yu2020financial}, predicting stock movements based on sentiments from financial news headlines and tweets \citep{sousa2019bert}, to summarizing financial reports \citep{la2020end}. Prefix tuning can be applied to many tasks with fewer parameters and much less memory consumption.

However, in the real world, the data might be noisy, especially in the case of chatbots and social media data where misspellings, typographical errors, and out-of-vocabulary words occur frequently. Recent studies have investigated the robustness of finetuning language models such as \citet{rychalska2019models},  \citet{jin2020bert}, \citet{aspillaga2020stress},  \citet{sun2020adv} and \citet{srivastava2020noisy}, and found that finetuning is not robust to noisy texts. 

To the best of our knowledge, there have been no studies that explore the robustness of prefix tuning that reflect real-life scenarios and compare it with finetuning to identify the more robust method. Our work corrupts the financial phrasebank dataset \citep{malo2014good}, using various text corruption methods such as keyboard errors (typos), inserting random characters, deleting random words, replacing characters with OCR alternatives and replacing words with antonyms by varying percentages in each sentence. The corrupted dataset is used with two widely used pre-trained models, BERT-base \citep{devlin2018bert} and RoBERTa-large \citep{liu2019roberta}, under both prefix tuning and fine-tuning, to compare their performance at different noise levels. In addition, we evaluate the performance on a Kaggle Stock Market Tweets dataset \cite{chaudhary_2020}, which is a real-life noisy dataset. 
With our experiments, we show that fine-tuning is more robust than prefix tuning in most setups. Fine-tuning updates the weights based on the downstream task and the dataset, and because of this, it can adapt to the noise, whereas prefix tuning uses the pre-trained model without updating the weights which limits the model from learning task-oriented information when facing noisy data. In summary, the contributions of this paper are three-fold.

\begin{itemize}
  \item To the best of our knowledge, this is among the first efforts in exploring the robustness of prefix tuning when facing noisy data, particularly noisy financial data.
  \item We use a comprehensive set of corrupted data and show that fine-tuning is more robust to noise compared to prefix tuning. The latter has also shown to have high variances in F1 scores.
  \item We provide detailed results at different levels of noise. With that, we advocate that caution should be carefully taken when practitioners apply state-of-the-art prefix tuning methods to noisy data. We hope our work will set baselines for further studies along this line. 
\end{itemize}

\section{Related Work}

\subsection{Sentiment Analysis in Financial Text}

Sentiment analysis is the process of understanding the sentiments from textual data \citep{liu2012sentiment}. Sentiment analysis in finance tries to achieve a different objective when compared to general sentiment analysis. Financial sentiment analysis aims to predict the stock movement or impact on stock price based on the sentiments of news headlines and news articles \citep{li2014news}. \citet{loughran2016textual} provide a survey of the machine learning approaches used to predict the sentiments in financial data.  With the introduction of transformer-based language models like BERT \citep{devlin2018bert}, several attempts have been made to predict the sentiments using the pre-trained BERT models trained on large text corpora.
\citet{araci2019finbert} introduced FinBERT, where the BERT model was pre-trained on a large financial corpus and it achieved state-of-the-art results in financial sentiment analysis. \citet{zhao2021bert} use RoBERTa \citep{liu2019roberta}, an optimized version of BERT to predict the sentiment of online financial texts generated on social media. 

\subsection{Robustness of Pretrained Language Models}

Several attempts have been made to test the robustness of popular transformer-based language models. \citet{rychalska2019models} test the robustness of ULMFiT \citep{howard2018universal} on various NLP tasks like QA, NLI, NER and Sentiment Analysis. The authors found that the high-performing language models are not robust to various corruption methods like removing articles, removing characters from words, misspellings, etc. \citet{jin2020bert} introduced a technique called TEXTFOOLER to generate adversarial texts. The authors successfully attacked BERT and significantly reduced the accuracy of BERT on text classification tasks. \citet{aspillaga2020stress} compared the robustness of RoBERTa, BERT and XLNET \citep{yang2019xlnet} with recurrent neural network models and found that RoBERTa, BERT and XLNET are more robust than recurrent neural networks but they are still not fully immune to the attacks and their robustness can be improved. \citet{sun2020adv} performed a detailed study on the robustness of BERT, especially concerning mistyped words (keyboard typos) and found that typos in informative words affect the performance of the BERT to a greater extent than typos in other words. \citet{srivastava2020noisy} analyzed the robustness of BERT to noise (spelling mistakes and typos) on sentiment analysis and textual similarity. The authors discovered that BERT's performance had significantly declined in the presence of noise in the text.


\begin{figure*}[ht]
     \centering
         \includegraphics[width=\textwidth]{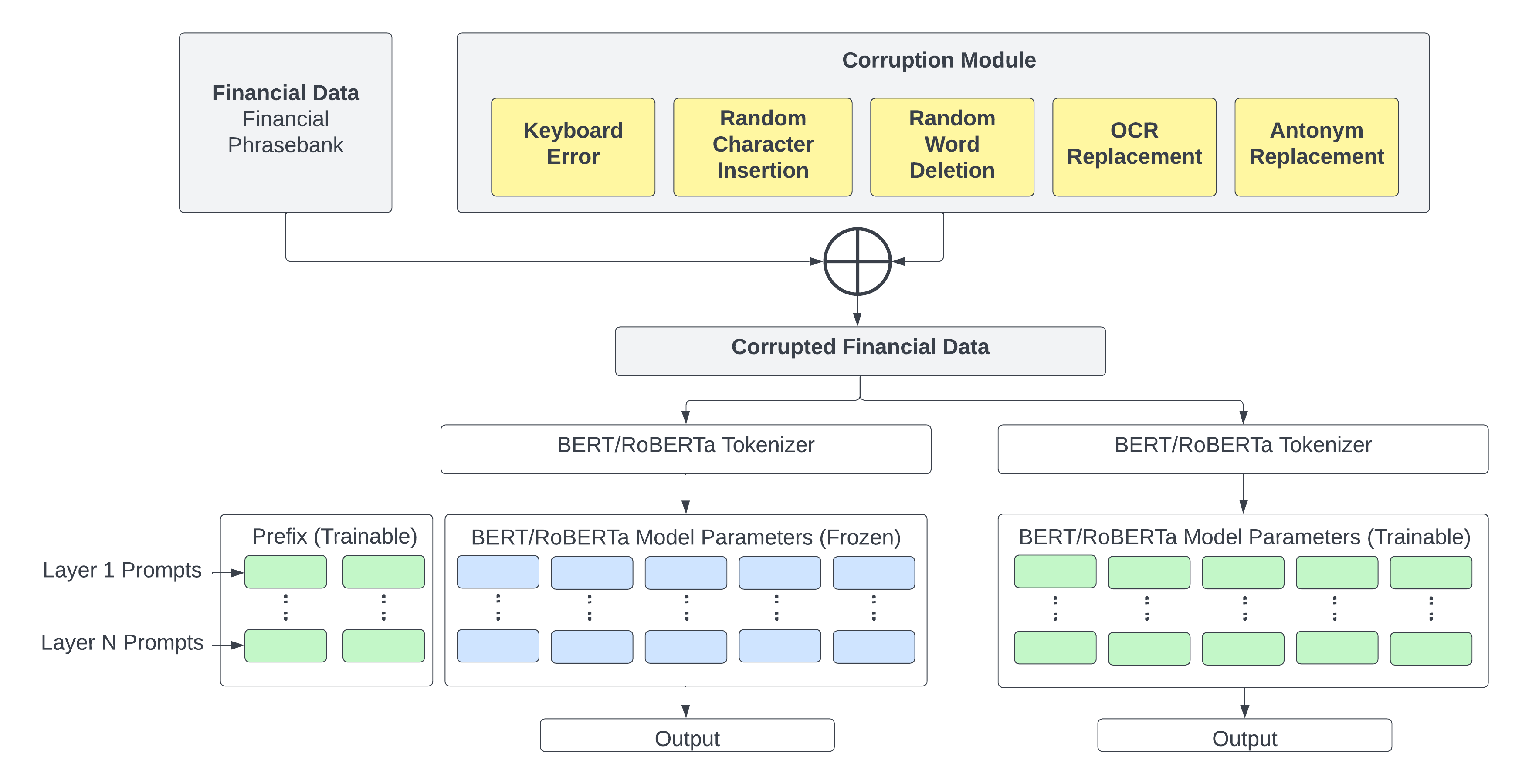}
        \centering
        \caption{Overview of prefix tuning and fine-tuning methodologies. Green boxes represent trainable parameters and blue boxes represent frozen parameters.}
        \label{fig:approach}
\end{figure*}


Prefix tuning freezes the parameters of the language model and updates the prefix vector for downstream tasks. \citet{yang2022robust} used the GPT-2 \citep{radford2019language} model to evaluate the robustness of prefix tuning to various textual adversarial attacks, but the attacks do not resemble the noise presented in real-world data. The authors did not compare the robustness of prefix tuning and fine-tuning and did not study which training methodology is more robust. 


\begin{table}
\caption{Train-validation-test split for the financial phrasebank 50\% agreement level dataset}
\begin{tabular}{m{1.5cm}m{1.5cm}m{1.5cm}m{1.5cm}m{1cm}}
\hline
\textbf{Label}
& {\textbf{Train Set}}& {\textbf{Validation Set}}& {\textbf{Test Set}}\\
\hline
Neutral & 2011 & 431 & 431\\
Positive &	954	& 204 &	205\\
Negative &	423	& 91 &	90\\
Total &	3388 & 726 & 726\\ \hline
\end{tabular}
\label{50_agree}
\end{table}

\begin{table}
\caption{Train-validation-test split for the financial phrasebank 100\% agreement level dataset}
\centering
\begin{tabular}{m{1.5cm}m{1.5cm}m{1.5cm}m{1.5cm}m{1cm}}
\hline
\textbf{Label}
& {\centering\textbf{Train Set}}& {\centering\textbf{Validation Set}}& {\centering\textbf{Test Set}}\\
\hline
Neutral & 973 & 209 & 209\\
Positive &	399	& 85 &	86 \\
Negative &	212	& 46 &	45\\
Total &	1584 & 340 & 340\\ \hline
\end{tabular}
\label{100_agree}
\end{table}

\begin{table}
\caption{Train-validation-test split for the Kaggle Stock Market Tweets dataset}
\centering
\begin{tabular}{m{1.5cm}m{1.5cm}m{1.5cm}m{1.5cm}m{1cm}}
\hline
\textbf{Label}
& {\centering\textbf{Train Set}}& {\centering\textbf{Validation Set}}& {\centering\textbf{Test Set}}\\
\hline
Positive &	2577 & 553 & 552\\
Negative &	1470 & 315  & 315 \\
Total &	4047 & 868 & 867\\ \hline
\end{tabular}
\label{kaggle_data}
\end{table}

\section{Approach}

\begin{table*}
\centering
\caption{
Corruption methods with an example
}
\begin{tabular}{m{4cm}m{11cm}}
\hline
\textbf{Corruption Method} & {\textbf{Example}}\\
\hline
Original Sentence & In Finland's Hobby Hall's sales decreased by 10\% , and international sales fell by 19\% . \\
Keyboard Error & In cinland' s Hubby Hall' s sales decreased by 10\% , and international saleW fell by 19\%. \\
Random Character Insertion & In FrinDIa*nd' s HZobJb\#y Hall's sales decreased by 10\% , and international sales fell by 19\% . \\
Random Word Deletion & In Finland' s Hobby Hall' s decreased 10\% , and international fell by 19\%. \\
OCR Replacement &In Finland' s H066y Ha11's sa1es decreased by 10\% , and national sales decreased by 19\%. \\
Antonym Replacement & In Finland' s Hobby Hall' s sales increase by 10\% , and national sales increase by 19\%. \\
\hline
\end{tabular}

\label{corruption-methods-examples}
\end{table*}

Figure~\ref{fig:approach} shows the overview of the approach used in this paper. The clean financial dataset is corrupted using the corruption module represented by yellow boxes in Figure~\ref{fig:approach}, containing various corruption methods. The corruption module is explained in section \ref{corruption_module}. The corrupted financial dataset is fed into two state-of-the-art pre-trained models, BERT-base and RoBERTa-large (refer to section \ref{pt_ft}) using both prefix tuning and fine-tuning. Figure~\ref{fig:approach} also shows the difference between the traditional fine-tuning method and prefix tuning respectively, where blue boxes represent the frozen parameters and green boxes represent the trainable parameters.

\subsection{Corruption Module}
\label{corruption_module}

The corruption module consists of 5 text corruption methods which closely replicate the noise found in real-world data. This module is used to corrupt the clean financial dataset and the corrupted dataset is used to evaluate the performance of the models. The following are the various text corruption methods used in the corruption module. Table~\ref{corruption-methods-examples} shows an example for each corruption method. The nlpaug \citep{ma2019nlpaug} library is used for generating the various corruption methods.

   \paragraph{Keyboard Error (QWERTY)} Simulates typing mistakes made while using a QWERTY-type keyboard.
   \paragraph{Random Character Insertion (ChIns)} Inserts random characters into a word in a sentence.
   \paragraph{Random Word Deletion (WdDel)} Randomly deletes a word from the sentence.
   \paragraph{OCR Replacement (OCR)} Replaces the characters in the word with their OCR equivalents, e.g., stock can be replaced as st0ck (here an alphabet, o, is replaced with the number zero, 0)
   \paragraph{Antonym Replacement (AntRep)} Replaces the words with their antonyms (opposite meaning) in the sentence.

\subsection{Prefix Tuning and Fine Tuning in Noisy Data}
\label{pt_ft}

\paragraph{Noisy Data Analysis} When the models BERT and RoBERTa encounter a word that is not in their vocabulary, the models try to break down the word to see whether any of its subwords are present in their vocabulary. For example, if BERT has the word `play' in its vocabulary and when it encounters `playing' it will tokenize the word as ```play' + `\#ing'''. If any word is not present in the vocabulary even after breaking it down, BERT assigns the unknown token (<UNK>) to that word. Table~\ref{tokenizer} shows how BERT and RoBERTa tokenize the normal and the corrupted word. From Table~\ref{tokenizer} we can understand how the corrupted word affects the BERT tokenizer and prevents it from learning the word's original meaning resulting in a drop in performance.

The process of prefix tuning and fine-tuning updating the weights is based on the downstream task and the dataset. In prefix tuning, most of the weights are not updated based on the downstream task. Since both the training and the validation sets are corrupted, in fine-tuning, the weights of the model have been updated based on the noisy datasets and contain more dataset-specific information than the prefix-tuned model. This enables the fine-tuned model to adjust to the noisy scenarios better than the prefix-tuned models. Evaluations of our intuition for prefix tuning and fine-tuning in noisy data can be found in Section~\ref{sec:experiments}.

\begin{table*}
\centering
\caption{
Tokenization of corruption variants for the word `stock'
}
\begin{tabular}{cccc}
\hline
\multicolumn{1}{c}{} & \multicolumn{1}{c}{} & \multicolumn{2}{c}{\textbf{Tokenized Word}} \\
\textbf{Corruption Method} & \textbf{Corrupted Word} & {\textbf{BERT}}  & {\textbf{RoBERTa}} \\
\hline
No Corruption & `stock' & [`stock'] & [`stock'] \\
Keyboard error & `srosk' & [`s', `\#\#ros', `\#\#k'] & [`s', `ros', `k'] \\
Random character insertion & `sto*rck' & [`s', `\#\#to', `*', `r', `\#\#ck'] & [`st', `o', `*', `r', `ck'] \\
OCR replacement & `st0ck' & [`s', `\#\#t', `\#\#0', `\#\#ck'] & [`st', `0', `ck'] \\
\hline
\end{tabular}
\label{tokenizer}
\end{table*}

\section{Experiments}
\label{sec:experiments}

\subsection{Financial Tasks}

Two financial tasks are used to evaluate the performance of prefix tuning. The first task is the sentiment analysis of the Financial Phrasebank dataset \citep{malo2014good}, which is the main dataset used to compare the performance and evaluate the robustness of both prefix tuning and fine-tuning. The second task is the sentiment analysis of the Twitter Stockmarket dataset from Kaggle, \citet{chaudhary_2020}, which is also used to evaluate the performance of prefix tuning and fine-tuning. 

\paragraph{Financial Phrasebank} The Financial Phrasebank dataset \citep{malo2014good}, consists of 4840 sentences from financial news articles and the sentences were manually labelled as positive, negative or neutral by 16 annotators with backgrounds in finance and business. The annotators labelled the sentences depending on whether the information from the sentence had a positive, negative or no impact on the stock prices of the company mentioned in the sentence. It is an imbalanced dataset with 1363 positive sentences, 604 negative sentences and 2873 neutral sentences. In addition to it, depending on the agreement level among the annotators on the polarity of the sentence, the dataset was classified into 50\%, 66\%, 75\% and 100\% agreement levels. For example, 50\% annotator agreement means more than 50\% of the annotators agreed and selected the same polarity for a particular sentence. This paper uses the financial phrasebank dataset with 50\% annotator agreement level (4840 sentences) to run the experiments on estimating the robustness of prefix tuning and the 100\% agreement level (2262 sentences) to compare the performance. The dataset was split into the train, validation and test sets for the experiments with a 70-15-15 split (stratified split) giving rise to 3388 training sentences, 726 validation sentences and 726 test sentences in the 50\% agreement level and  1582 training sentences, 340 validation sentences and 340 test sentences in the 100\% agreement level dataset. Table~\ref{50_agree} shows the split up of the 50\% agreement level dataset and Table~\ref{100_agree} shows the split up of the 100\% agreement level dataset.

\paragraph{Kaggle Stock Market Tweets} The Stock Market tweets dataset is from Kaggle, \citet{chaudhary_2020}. The reason for selecting this dataset is to evaluate the performance of prefix tuning and fine-tuning on a real-world noisy data. This dataset contains tweets from Twitter consisting of information about the stocks of multiple companies and the tweets are labelled as either positive or negative based on the sentiment associated with each tweet. This dataset is from Kaggle and it is not from a renowned journal and the authenticity cannot be validated. The dataset consists of 2106 negative tweets and 3685 positive tweets. The dataset was split into the train, validation and test sets with a 70-15-15 split giving rise to 4047 training sentences, 868 validation sentences and 867 test sentences. Table~\ref{kaggle_data} shows the split up of the Kaggle Stock Market dataset.

\subsection{Setup}

\begin{table*}
\centering
\caption{
Results for the uncorrupted version of the datasets for the BERT-base model
}
\begin{tabular}{crrrr}
\hline
\multicolumn{1}{c}{} & \multicolumn{2}{c}{\textbf{Prefix Tuning}} & \multicolumn{2}{c}{\textbf{Fine Tuning}}  \\ 
\textbf{Dataset} & {\textbf{Acc.(\%})}  & {\textbf{F1(\%)}}  & {\textbf{Acc.(\%)}}  & {\textbf{F1(\%)}} \\
\hline
Financial Phrasebank - All agree &  97.35 &  97.01 & 96.17 & 96.80  \\
Financial Phrasebank - More than 50\% agree & 86.91 & 85.55 & 86.09 & 85.48  \\
Kaggle Stock Market Tweets & 79.60 & 77.74 & 80.41 & 78.96  \\
\hline
\end{tabular}

\label{clean-datasets-bert}
\end{table*}

\begin{table*}
\centering
\caption{
Results for the uncorrupted version of the datasets for the RoBERTa-large model
}
\begin{tabular}{crrrr}
\hline
\multicolumn{1}{c}{} & \multicolumn{2}{c}{\textbf{Prefix Tuning}} & \multicolumn{2}{c}{\textbf{Fine Tuning}}  \\ 
\textbf{Dataset} & {\textbf{Acc.(\%})}  & {\textbf{F1(\%)}}  & {\textbf{Acc.(\%)}}  & {\textbf{F1(\%)}} \\
\hline
Financial Phrasebank - All agree &  98.24 & 98.09 & 98.53 & 98.35  \\
Financial Phrasebank - More than 50\% agree & 87.60 & 87.25 & 88.15 & 87.45  \\
Kaggle Stock Market Tweets & 81.79 & 79.61 & 82.71 & 80.61  \\
\hline
\end{tabular}

\label{clean-datasets-roberta}
\end{table*}

\begin{table*}
\centering
\caption{
Financial Phrasebank results for various text corruption methods for both the BERT-base and the RoBERTa-large model
}
\begin{tabular}{cr|rrrr|rrrr}
\hline
&&\multicolumn{4}{c|}{\textbf{BERT-base}} & \multicolumn{4}{c}{\textbf{RoBERTa-large}}  \\ 
\multicolumn{1}{c}{} & \multicolumn{1}{c|}{\textbf{Cor.}} & \multicolumn{2}{c}{\textbf{Prefix Tuning}} & \multicolumn{2}{c|}{\textbf{Fine Tuning}} & \multicolumn{2}{c}{\textbf{Prefix Tuning}} & \multicolumn{2}{c}{\textbf{Fine Tuning}} \\ 
\textbf{Method} & {\textbf{(\%)}}  & {\textbf{Acc.}}  & {\textbf{F1}}  & {\textbf{Acc.}}  & {\textbf{F1}} & {\textbf{Acc.}}  & {\textbf{F1}}  & {\textbf{Acc.}}  & {\textbf{F1}}\\
\hline
None & \centering- & 86.91 & 85.55 & 86.09 & 85.48 &  87.60 & 87.25 & 88.15 & 87.45\\
\hline
\multirow{5}{*}{Qwerty} & \centering10 & 0.47 & -0.47  & -0.16 & -0.5 & -0.78 & -1.60 & -0.94 & -1.63\\
& \centering20 & -0.96 & -0.44 & -0.01 & -0.43 & -1.40 & -1.43 & -1.08 & -0.82\\
& \centering30 & -0.16 & -0.68 & 0.15 & -1.09 & -4.37  & -5.50 & -3.26 & -4.64\\
& \centering40 & -1.58 & -2.50 & -0.81 & -1.12 & -6.21 & -8.85 & -3.56 & -4.55\\
& \centering50 & -6.34 & -8.87 & -3.04 & -5.11 & -6.21 & -9.04 & -3.57 & -4.80\\
\hline
\multirow{5}{*}{ChIns} & \centering10 & -1.10 & -1.27 & -0.65 & -1.31 & -0.46 & -1.58 & -0.46 & -1.84\\
& \centering20 & -3.01 & -3.53 & -0.96 & -1.37 & -2.80 & -3.30 & -1.86 & -3.09\\
& \centering30 & -3.33 & -4.57 & -0.96 & -2.40 & -2.78 & -3.81  & -3.73 & -5.15\\
& \centering40 & -3.01 & -4.63 & -3.68 & -4.22 & -2.64 & -4.41 & -4.04 & -4.24\\
& \centering50 & -5.38 & -6.63 & -2.89 & -4.73 & -5.56 & -8.31 & -5.13 & -7.40\\
\hline
\multirow{5}{*}{WdDel} & \centering10 & -0.63 & -0.04 & 0.15 & -1.42 & -0.94 & -1.45  & -1.40  & -1.26\\
& \centering20 & -0.96 & -0.98 & -0.82 & -1.70 & -1.70 & -2.48 & -1.56 & -2.27\\
& \centering30 & -1.90 & -2.49 & -1.29 & -2.06 & -4.04 & -3.75 & -2.80 & -3.69\\
& \centering40 & -3.01 & -4.63 & -0.80 & -2.29 & -1.70 & -2.05  & -0.94 & -1.74\\
& \centering50 & -5.38 & -6.63 & -3.21 & -3.11 & -2.02 & -2.86 & -2.02 & -2.29\\
\hline
\multirow{5}{*}{OCR} & \centering10 & -0.63 & -0.28 & -1.13 & -1.04 & -0.78 & -1.31 & -0.78 & -1.89\\
& \centering20 & -0.79 & -1.30 & -0.01 & -0.27 & -0.94 & -1.42 & -0.78 & -1.67\\
& \centering30 & -2.53 & -2.93 & -2.73 & -2.69 & -2.48 & -3.01 & -3.57 & -4.04\\
& \centering40 & -2.53 & -2.93 & -2.73 & -2.69 & -4.66 & -5.08 & -2.80 & -4.34\\
& \centering50 & -7.44 & -11.30 & -10.73 & -10.45 & -5.13 & -8.59 & -4.19 & -6.62\\
\hline
AntRep & \centering- & -12.36 & -25.55 & -14.25 & -27.41 & -14.13 & -28.20 & -16.78 & -30.05\\
\hline
\end{tabular}

\label{corruption-methods}
\end{table*}

\paragraph{Corruption Strategy} The clean versions of the financial phrasebank dataset, 100\% agreement level and 50\% agreement level, are used to evaluate the performance of prefix tuning and fine-tuning on both BERT-base and RoBERTa-large models to establish the baseline performance levels. To test the robustness of prefix tuning and find out which one between prefix tuning and fine-tuning is more robust to the noisy text, the train and validation sets of the financial phrasebank dataset (50\% agreement level) are corrupted by various text corruption methods. The reason for corrupting the train and validation sets is that it is difficult to find large-scale high-quality training data, especially with respect to chatbots and social media texts in an industrial setting. In general, test data is smaller in size compared to the training data and can be manually cleaned before feeding into the model. Due to this, the training and validation sets have been corrupted. For each corruption method, the sentences are corrupted by 10\%, 20\%, 30\%, 40\% and 50\% corruption levels. Each corruption level represents the percentage of corrupted words in a sentence. For example, 10\% corruption level means 10\% of the words in the sentence are corrupted. For antonym replacement, all the words which have antonyms in the nlpaug \citep{ma2019nlpaug} library are replaced with antonyms and there are no varying corruption levels for this particular corruption method.

\paragraph{Implementation Details} After corrupting the dataset using the above-mentioned corruption strategy, we conduct the experiments on two models, BERT base and RoBERTa large. The BERT base fine-tuned model has 108,312,579 trainable parameters while the prefix-tuned model has 370,947 trainable parameters for 30 epochs for the financial phrasebank dataset. Similarly, the RoBERTa large fine-tuned model has 355,362,819 trainable parameters while the prefix-tuned model has 986,115 trainable parameters for 30 epochs for the financial phrasebank dataset. More information about the implementation details can be found in Appendix \ref{appendix:implementation_details} for replication.

\paragraph{Evaluation Metrics} The F1 score and accuracy are selected as the metrics for the evaluation of the experiments. The F1 score is used as the main metric for comparison since the financial phrasebank is an imbalanced dataset with 3 classes, positive, negative and neutral.

\subsection{Results}

\begin{table*}
\centering
\caption{
Mean and Variance of F1 scores for the BERT-base model for 50\% noise level
}
\begin{tabular}{crrrr}
\hline
\multicolumn{1}{c}{} & \multicolumn{2}{c}{\textbf{Prefix Tuning}} & \multicolumn{2}{c}{\textbf{Fine Tuning}}  \\ 
\textbf{Corruption Method} & {\textbf{ Mean (F1\%})}  & {\textbf{Variance}}  & {\textbf{Mean (F1\%)}}  & {\textbf{Variance}} \\
\hline
No Corruption & 85.48 & 0.16 & 85.48 & 0.13  \\
Keyboard error & 80.57 & 5.66 & 82.00 & 1.15  \\
Random character insertion & 80.84 & 0.72 & 81.97 & 0.86  \\
Random word deletion & 81.98 & 0.10 & 82.50 & 0.13  \\
OCR replacement & 75.77 & 3.64 & 77.49 & 0.86  \\
Antonym replacement & 64.05 & 1.94 & 62.23 & 0.06  \\
\hline
\end{tabular}

\label{mean-variance}
\end{table*}

\begin{table*}
\caption{
Predicted labels for BERT-base OCR replacement 50\% corruption level in cases where fine-tuning predicted the correct labels and prefix tuning predicted the wrong labels}
\centering
\newcolumntype{U}{>{\centering\arraybackslash}m{7cm}}
\newcolumntype{V}{>{\centering\arraybackslash}m{2cm}}
\begin{tabular}{m{7cm}m{2.4cm}m{2.4cm}m{2.4cm}}
\hline
\centering{\textbf{Sentence}} & \textbf{True Label} & \multicolumn{2}{c}{\textbf{Predicted Label}} \\
 &  & \textbf{Prefix Tuning}& \textbf{Fine Tuning} \\
\hline
The amending of the proposal simplifies the proposed plan and increases the incentive for key employees to stay in the Company & Positive & Neutral & Positive \\
The company 's net sales in 2009 totalled MEUR 307.8 with an operating margin of 13.5 per cent & Neutral & Positive & Neutral \\
The move was triggered by weak demand for forestry equipment and the uncertain market situation & Negative & Neutral & Negative \\
\hline
\end{tabular}
\label{ft_correct}
\end{table*}

\paragraph{Clean Baselines} Table~\ref{clean-datasets-bert} and Table~\ref{clean-datasets-roberta} show the performance of both models on the clean versions of the financial phrasebank dataset and the noisy Kaggle stock market tweets dataset (uncorrupted). Both prefix tuning and fine-tuning achieve comparable performance in both clean versions of the financial phrasebank dataset (all agree and more than 50\% agree). In the noisy tweets dataset, fine-tuning performs better than prefix tuning in both models. The Bert-base finetuning method achieves an F1 score of 78.96\% which is greater than prefix tuning (77.74\%) by 1.22 point F1 score. Similarly, RoBERTa fine-tuning method achieves an F1 score of 80.61\% which is greater than prefix tuning (79.61\%) by 1 point F1 score.

 
\paragraph{Corruption Results} Table~\ref{corruption-methods} shows the change in the baseline scores of prefix tuning and fine-tuning on different corruption methods for BERT-base and RoBERTa-large respectively. The performance of both fine-tuning and prefix tuning drops as the noise level increases. 
Overall, finetuning performs better than prefix tuning in all the corruption methods except for antonym replacement. Even though the difference in F1 scores is very minimal for the lower percentage of noise like 10\% and 20\%, the difference becomes more predominant when the noise percentage in each sentence increases. This trend can be observed in both BERT-base and RoBERTa-large models. 

To further evaluate the validity of the results, the variance (how the F1 scores vary from mean F1 scores across various iterations) for 50\% noise level for all the corruption methods is measured. The experiments were repeated 5 times with reshuffled data for all the corruption methods to measure the mean and variance of F1 scores. Table~\ref{mean-variance} shows the mean and variance of F1 scores for the BERT-base model. It can be observed that the variance for prefix tuning is very high in two corruption methods, keyboard (qwerty) error and OCR replacement error. 

There is a significant drop in performance (more than 25\%) for antonym replacement. Fine tuning achieves an F1 score of 62.05\% whereas prefix tuning achieves an F1 score of 63.69\%. When compared to prefix tuning, the fine-tuned model achieves lower performance and it could be due to the following reason. The weights of the fine-tuned model are updated with the corrupted dataset containing antonyms instead of the original words. Since the model is trained to predict the opposite sentiment (sentences with antonyms), the performance drops significantly when evaluated on the test dataset. This results in the fine-tuned model being more adapted to the corrupted dataset and achieving lower performance when exposed to a clean test dataset whereas prefix tuning performs comparatively better.

Table~\ref{ft_correct} shows the predicted labels for BERT-base OCR replacement 50\% corruption level where fine-tuning predicted the correct labels and prefix tuning predicted the wrong labels. In most of the cases, the positive labels were incorrectly predicted as neutral, the neutral labels were incorrectly predicted as positive and the negative labels were incorrectly predicted as neutral. 


Another interesting observation is the minimal performance drop seen in the random word deletion corruption method even when 50\% of the words are deleted from the sentences. The performance drop in the F1 score for the BERT base model was 6.63\% for prefix tuning and 3.11\% for fine-tuning. Similarly, the performance drop in the F1 score for the RoBERTa large model was 2.86\% for prefix tuning and 2.29\% for fine-tuning. The main reason behind this could be the way BERT is trained. BERT uses masked language modelling where it masks the words at random by varying percentages and tries to predict the masked word based on the context. This might be the reason why there is no significant drop in performance even when deleting 50\% of the words since both BERT and RoBERTa are trained to handle the missing words in a sentence.

\section{Conclusion}

With the sizes of pre-trained models continuing to be significantly larger, lightweight models have become more important for many financial applications. However, the 
robustness of such models has not been well understood yet. In this paper, we explored the robustness of prefix tuning by corrupting the financial phrasebank dataset with various corruption methods, including keyboard (qwerty) error, random character insertion, OCR replacement, random word deletion and antonym replacement under varying noise levels at 10\%, 20\%, 30\%, 40\% and 50\%, as well as on the Kaggle stock market tweets, which is a real-world noisy dataset. We show that fine-tuning is more robust to noise than prefix tuning in most of the corruption methods. As the impact of noise is more significant along with increasing noise levels, prefix tuning shows a more significant decrease in performance compared to full fine-tuning. The variance of performance of prefix tuning is higher than that of fine-tuning for most corruption setups. Our study suggests that caution should be taken by practitioners when applying prefix tuning to noisy data. A solution to improving the robustness to reduce the impact of noise is desired and is our immediate future work.

\section{Limitations}

\noindent The words were randomly corrupted in a sentence with no emphasis on the word's context and no experiments were carried out to find out the importance of the corrupted word in the context of predicting the sentiment. Corrupting an important word may result in an increased drop in performance than corrupting a word which has minimal impact on the sentiment of a sentence. \citet{sun2020adv} have found that typos on informative words affect the performance of the BERT to a greater extent than typos in other words. Furthermore, the robustness was evaluated on the sentiment analysis task and it was not evaluated on other natural language processing tasks like question answering, named entity recognition and text summarization.

\bibliography{anthology,custom}
\bibliographystyle{acl_natbib}

\appendix
\section{Appendix}
\subsection{Implementation Details} \label{appendix:implementation_details}

The experiments were carried out on four Nvidia GeForce RTX 2080 GPU's for 30 epochs. The length of the prefix plays a significant role in prefix tuning. In \cite{liu2021p}, the authors have suggested that Natural Language Understanding (NLU) tasks prefer shorter prefix lengths and they have used a prefix length of 20 for sentiment classification to obtain the best performance. We have also used a prefix length of 20 to evaluate the performance of the models. The learning rate differs for each model and method. For prefix tuning, both BERT-base and RoBERTa-large models use a learning rate of 1e-2. For fine-tuning, BERT-base uses a learning rate of 2e-5 and RoBERTa-large used a learning rate of 2e-6. Furthermore, the 50\% noise level is selected for all the corruption methods and the variance is measured for both prefix tuning and fine-tuning for the BERT base model. The Kaggle Stock Market tweets dataset is also used to evaluate the performance of prefix tuning and fine-tuning on real-world noisy data (tweets) with the same set of hyperparameters as the financial phrasebank dataset.

\subsection{Experimental Results - Visualizations}\label{appendix:extra_results}

\begin{figure*}[ht]
     \centering
     \begin{subfigure}[b]{0.45\textwidth}
         \centering
         \includegraphics[width=\textwidth]{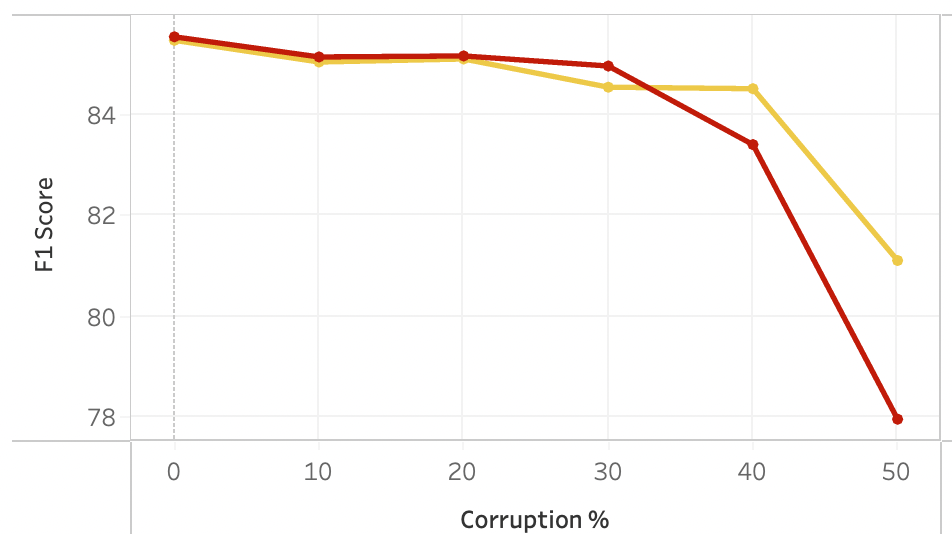}
         \caption{Keyboard Error}
         \label{fig:keyboard_small}
     \end{subfigure}
     \hfill
     \begin{subfigure}[b]{0.45\textwidth}
         \centering
         \includegraphics[width=\textwidth]{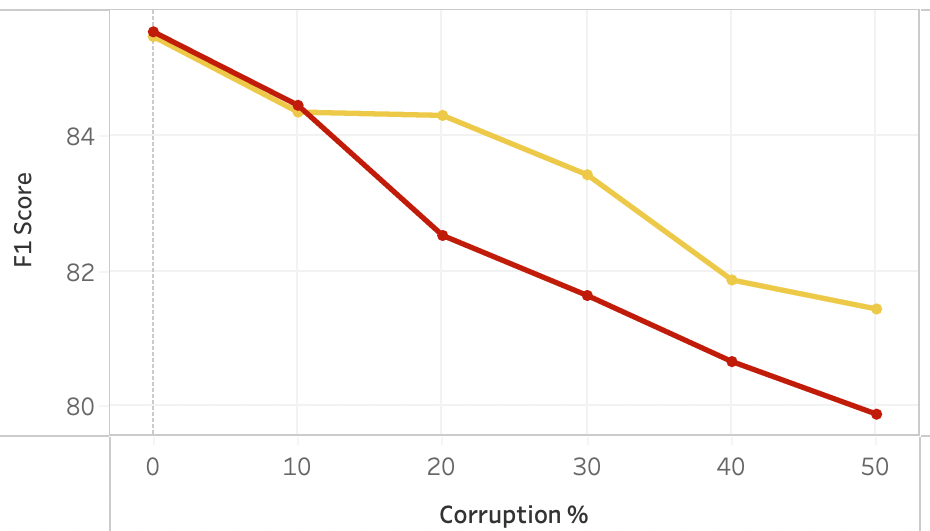}
         \caption{Random Char. Insertion}
         \label{fig:insertion_small}
     \end{subfigure}
     \hfill
     \begin{subfigure}[b]{0.45\textwidth}
         \centering
         \includegraphics[width=\textwidth]{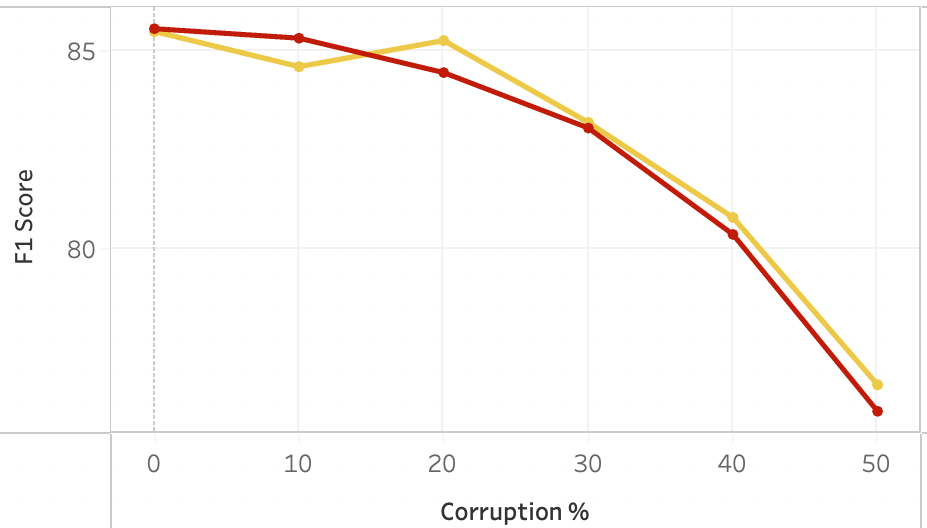}
         \caption{OCR Replacement}
         \label{fig:ocr_replacement}
     \end{subfigure}
     \hfill
     \begin{subfigure}[b]{0.45\textwidth}
         \centering
         \includegraphics[width=\textwidth]{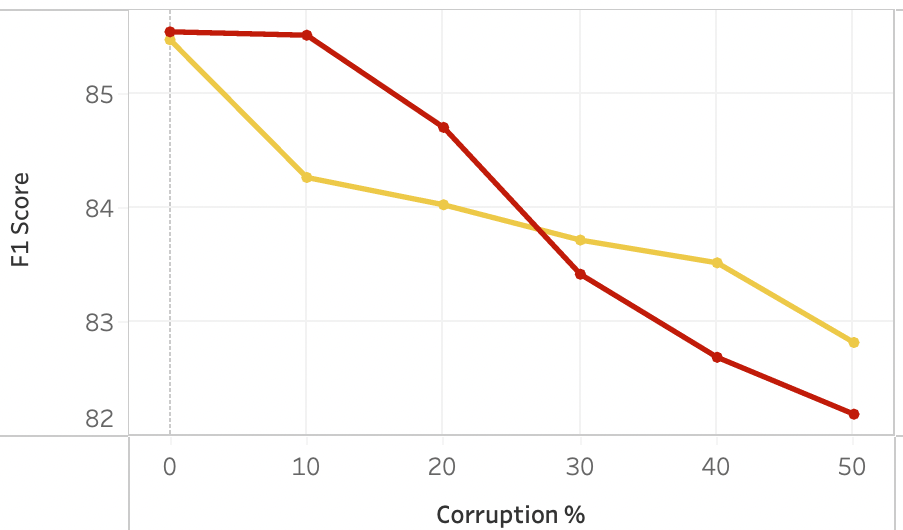}
         \caption{Random word Deletion}
         \label{fig:deletion_small}
     \end{subfigure}
        \centering
        \caption{Plot of F1 scores of BERT-base model for various corruption methods. Red line represents prefix tuning and yellow line represents fine-tuning.}
        \label{fig:bert base results}
\end{figure*}

\begin{figure*}[ht]
     \centering
     \begin{subfigure}[b]{0.45\textwidth}
         \centering
         \includegraphics[width=\textwidth]{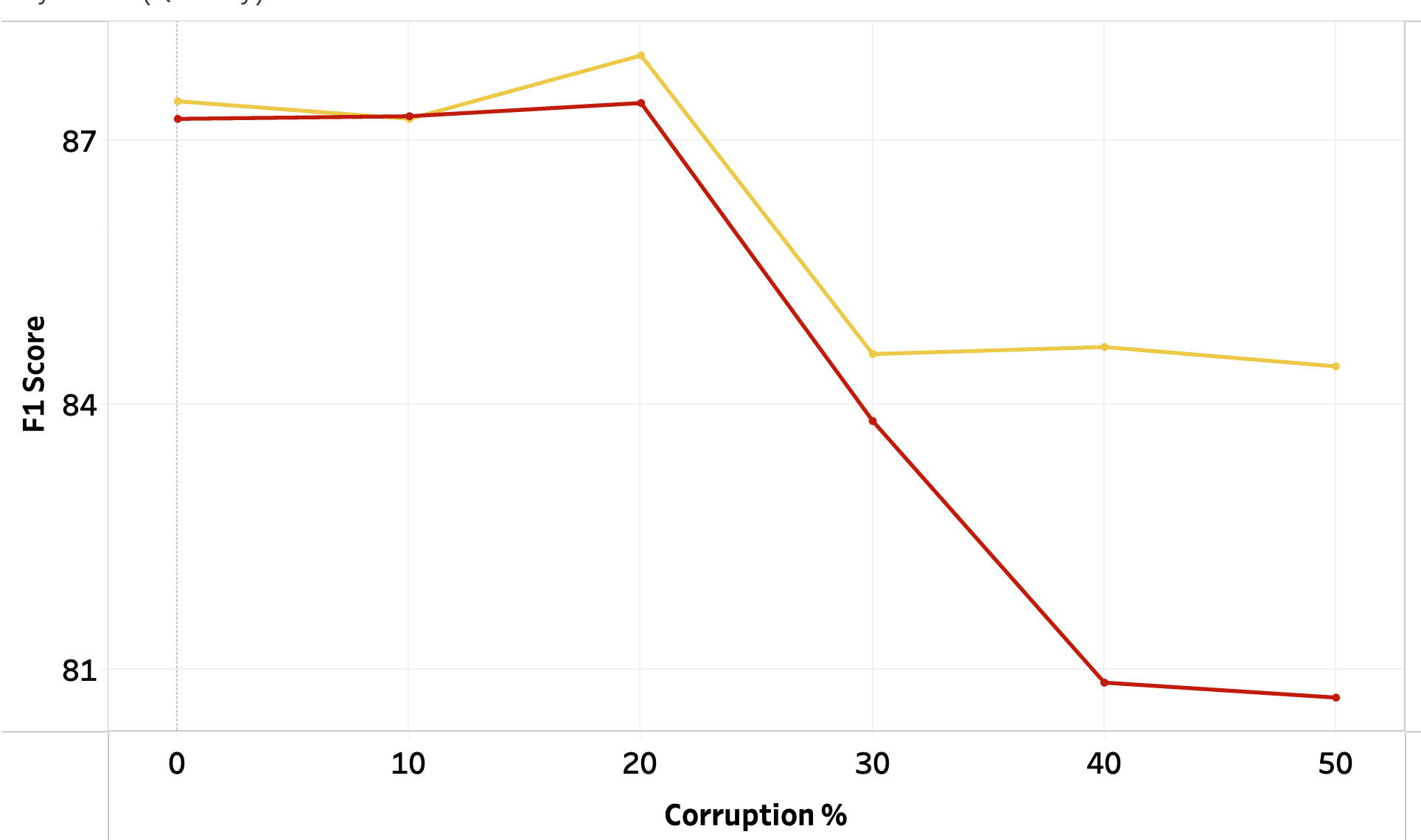}
         \caption{Keyboard Error}
         \label{fig:keyboard_large}
     \end{subfigure}
     \hfill
     \begin{subfigure}[b]{0.45\textwidth}
         \centering
         \includegraphics[width=\textwidth]{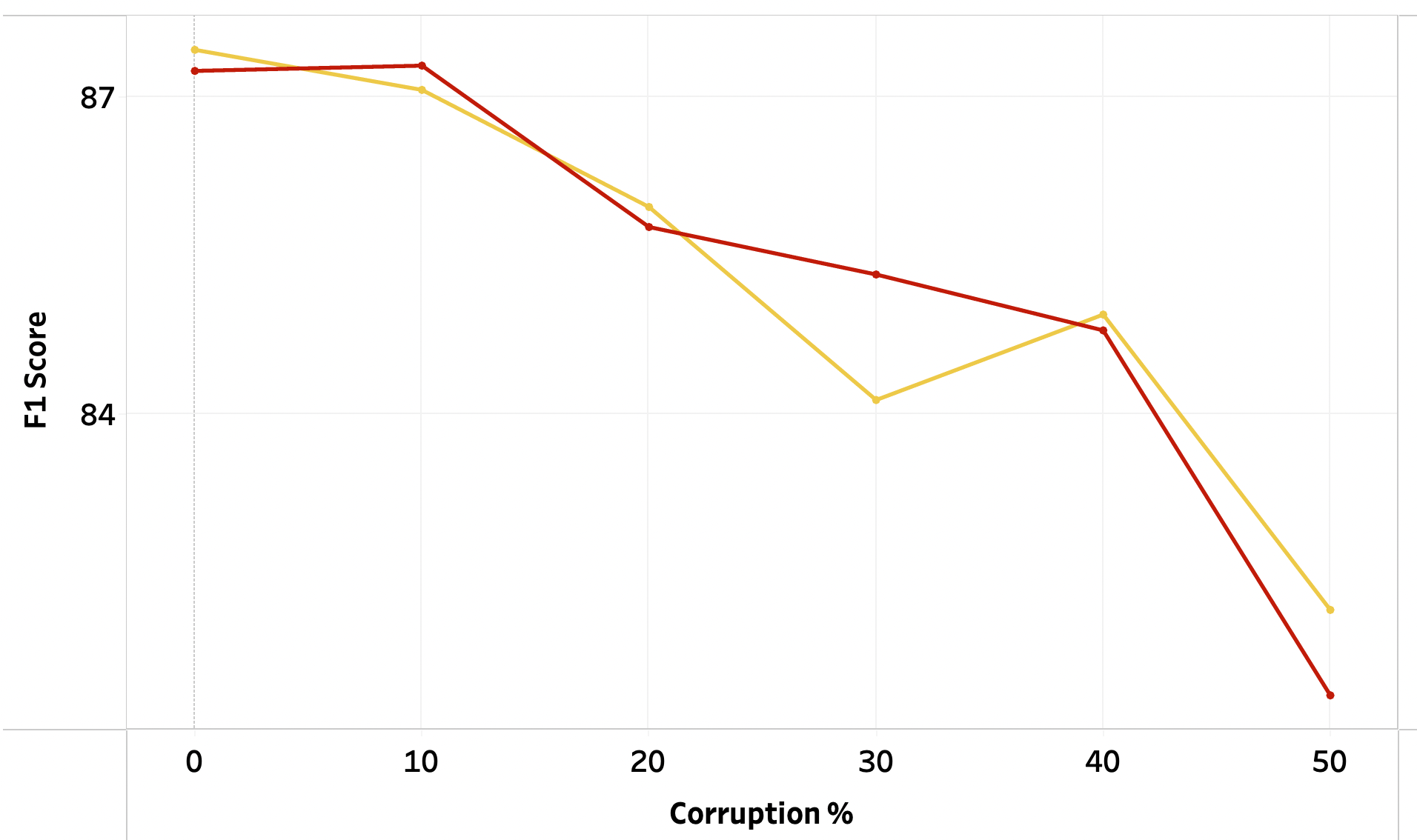}
         \caption{Random Char. Insertion}
         \label{fig:insertion_large}
     \end{subfigure}
     \hfill
     \begin{subfigure}[b]{0.45\textwidth}
         \centering
         \includegraphics[width=\textwidth]{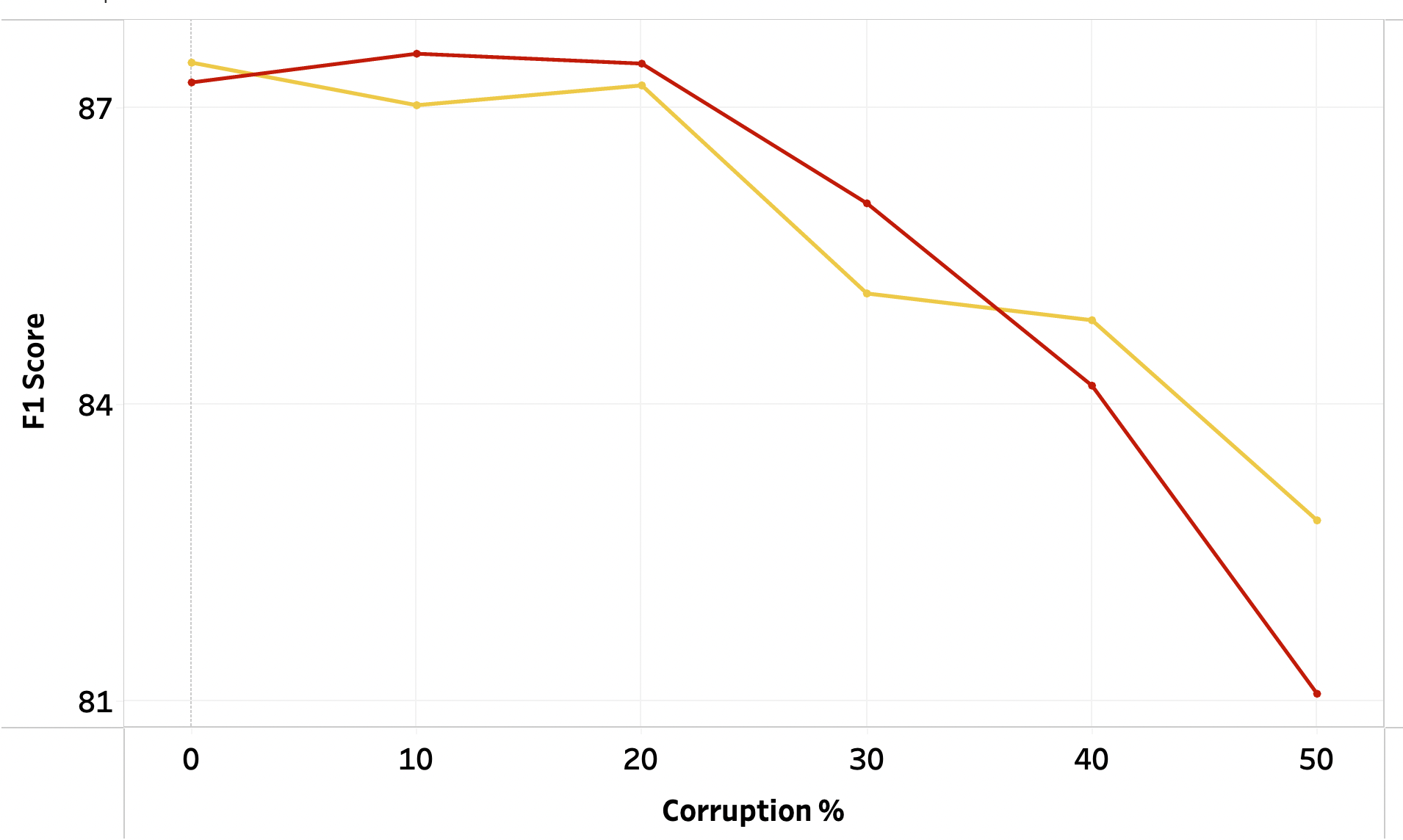}
         \caption{OCR Replacement}
         \label{fig:ocr_large}
     \end{subfigure}
     \hfill
     \begin{subfigure}[b]{0.45\textwidth}
         \centering
         \includegraphics[width=\textwidth]{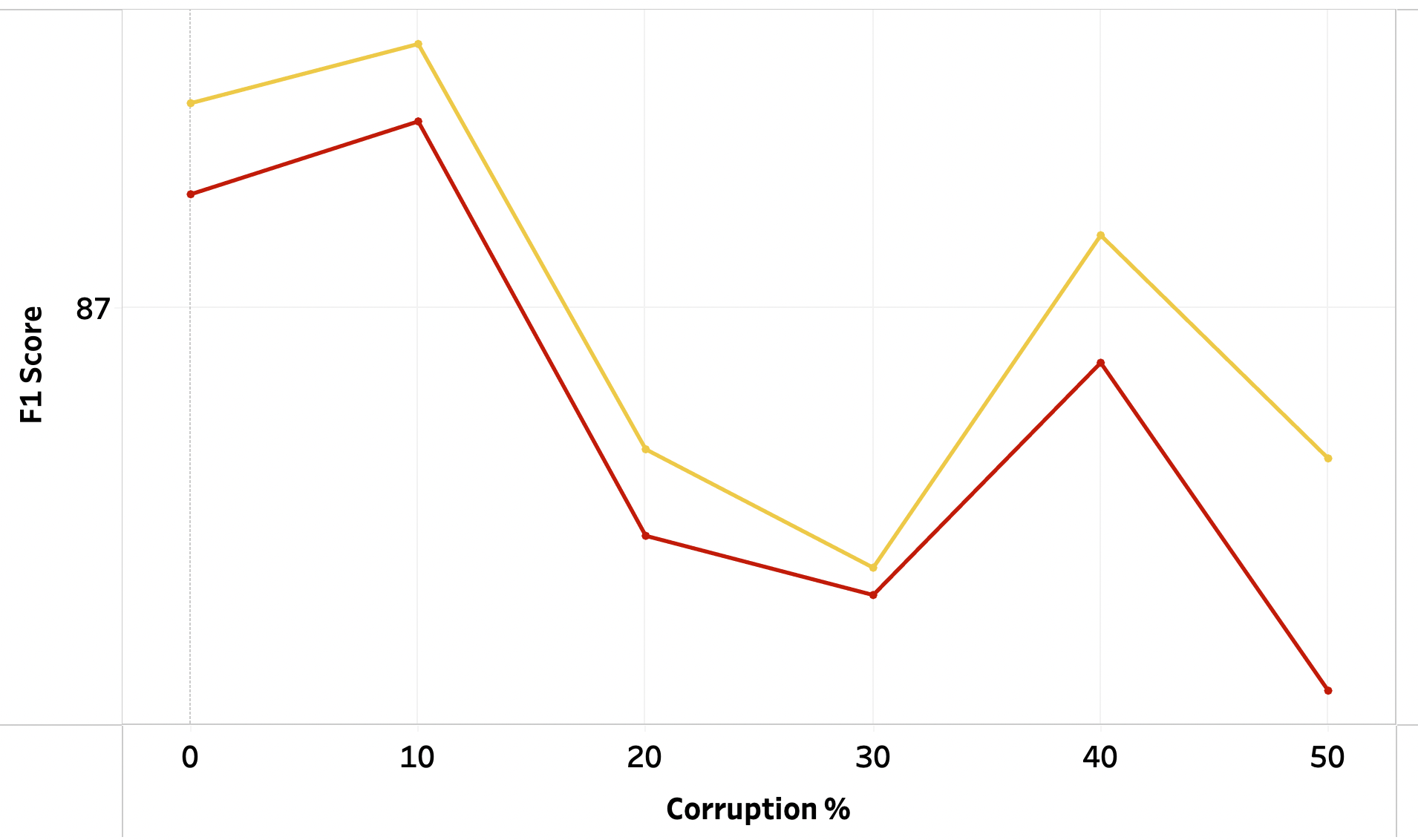}
         \caption{Random word Deletion}
         \label{fig:deletion_large}
     \end{subfigure}
        \centering
        \caption{Plot of F1 scores of RoBERTa-large model for various corruption methods. Red line represents prefix tuning and yellow line represents fine-tuning.}
        \label{fig:roberta large results}
\end{figure*}

\end{document}